\ifdefined\XeTeXversion\else\ifdefined\pdfoutput\pdfoutput=1\fi\fi

\documentclass[letterpaper, 10pt, conference]{ieeeconf}

\ifdefined\XeTeXversion
  \usepackage{newtxtext}
\else
  \usepackage[T1]{fontenc}
\fi
\usepackage{amssymb}
\usepackage{comment}
\usepackage{myconfig}

\IEEEoverridecommandlockouts

\newenvironment{IEEEkeywords}%
  {\vspace{0.5\baselineskip}\noindent\small\textbf{\textit{Index Terms}}---}%
  {\vspace{0.5\baselineskip}}

\title{\LARGE \bf
Tactile Memory with Soft Robot: Robust Object Insertion via Masked Encoding and Soft Wrist
}

\author{Tatsuya Kamijo$^{\dagger,1^*,2}$, Mai Nishimura$^{\dagger,1}$, Nodoka Shibasaki$^{1^*,3}$,
Jeremy Siburian$^{1^*,2}$, \\ Cristian C. Beltran-Hernandez$^{1}$, and Masashi Hamaya$^{1}$%
\thanks{$^{\dagger}$Equal contribution.}%
\thanks{$^{1}$ OMRON SINIC X Corporation, 5-24-5 Hongo, Bunkyo-ku, Tokyo, 113-0033, Japan (e-mail: {\tt\small masashi.hamaya@sinicx.com}).}
\thanks{$^{2}$ The University of Tokyo, 7-3-1 Hongo, Bunkyo-ku, Tokyo, 113-8654, Japan.}
\thanks{$^{3}$ Northeastern University, 360 Huntington Ave, Boston, MA, 02115, USA.}
\thanks{$^{*}$ Work done during internships at OMRON SINIC X.}
\thanks{This work was supported by JST, Japan (ACT-X Grant Number JPMJAX22AC, AIP Acceleration Research JPMJCR23U2, PRESTO JPMJPR24T9, JPMJPR2518), and JSPS KAKENHI Grant Number 25K00150.}
}

\begin{document}

\twocolumn[{%
\noindent
\copyright{} 2026 IEEE. Personal use of this material is permitted. Permission from IEEE must be obtained for all other uses, in any current or future media, including reprinting/republishing this material for advertising or promotional purposes, creating new collective works, for resale or redistribution to servers or lists, or reuse of any copyrighted component of this work in other works.\\

\noindent
\textbf{Accepted article:}\\
T. Kamijo, M. Nishimura, N. Shibasaki, J. Siburian, C. C. Beltran-Hernandez, and M. Hamaya, ``Tactile Memory with Soft Robot: Robust Object Insertion via Masked Encoding and Soft Wrist,'' \textit{IEEE Robotics and Automation Letters}, vol. 11, no. 7, pp. 7844--7851, July 2026. DOI: \href{https://doi.org/10.1109/LRA.2026.3692097}{10.1109/LRA.2026.3692097}.
}]
\thispagestyle{empty}
\pagenumbering{gobble}
\clearpage

\maketitle
\thispagestyle{empty}
\pagestyle{empty}

\begin{abstract}
Tactile memory, the ability to store and retrieve touch-based experience, is critical for contact-rich tasks such as key insertion under uncertainty. 
To replicate this capability, we
introduce Tactile Memory with Soft Robot (TaMeSo-bot), a system that integrates a soft wrist with tactile retrieval-based control to enable safe and robust manipulation. 
The soft wrist allows safe contact exploration during data collection, while tactile memory reuses past demonstrations via retrieval for flexible adaptation to unseen scenarios.
The core of this system is the Masked Tactile Trajectory Transformer (\mattt), which jointly models spatiotemporal interactions between robot actions, distributed \rev{tactile cues}, force-torque measurements, and proprioceptive signals.
Through masked token prediction, \mattt~learns rich spatiotemporal representations by inferring missing sensory information from context, autonomously extracting task-relevant features without explicit subtask segmentation.
We validate our approach on peg-in-hole tasks with diverse pegs and conditions in real-robot experiments. 
Our extensive evaluation demonstrates that \mattt~achieves higher success rates than the baselines over all conditions and shows remarkable capability to adapt to unseen pegs and conditions.
\end{abstract}


\section{Introduction}
\label{sec:introduction}

A tactile memory system, which stores and retrieves tactile information, is essential for daily activities~\cite{gallace2009cognitive}. For example, humans can insert a key even when the exact location of the lock is uncertain by exploring the surrounding area through gentle touch and relying on the tactile memory of how the key fits and inserts into the lock.
This ability to handle uncertainty through \rev{tactile cues} is a fundamental aspect of human robust manipulation.
While this process is trivial for humans, replicating such intuitive and adaptive behavior in robots remains a major challenge. 
This study explores a robotic system, {\bf Ta}ctile {\bf Me}mory with {\bf So}ft Ro{\bf bot} ({\bf TaMeSo-bot}), that integrates two key elements: soft robots, which enable stable contact, and a tactile memory capable of storing and retrieving tactile information in a database.

\begin{figure}[t]
    \centering
    \includegraphics[width=\linewidth]{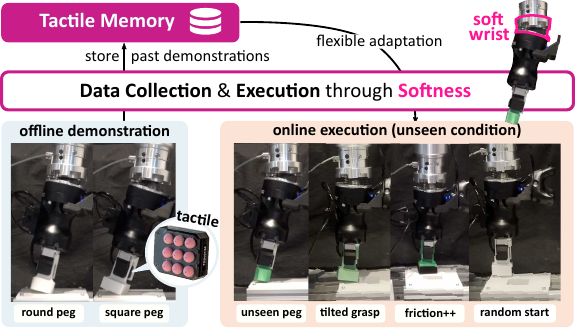}
    \caption{\textbf{TaMeSo-bot} uses softness (i.e., mechanical compliance) to 
safely collect tactile demonstrations (left) and flexibly adapt to unseen 
conditions (right) through retrieval from tactile memory. 
}
    \label{fig:concept}
\end{figure}

The goal of this research is to achieve robust peg-in-hole insertion, a widely studied yet challenging contact-rich manipulation task~\cite{suomalainen2022survey, jin2021contact}, through a tactile memory system.
As depicted in \cref{fig:concept},
robustness refers to exhibiting adaptive behavior across unseen pegs and conditions.
TaMeSo-bot enables flexible adaptation to novel scenarios by storing and retrieving past tactile demonstrations,
using a largely deformable soft wrist~\cite{von2020compact} for both safe data collection and robust execution.
Unlike prior approaches that rely on predefined contact states or contact-formation abstractions~\cite{johannsmeier2019framework, morgan2023towards}, 
our system requires no manual annotation of subtasks such as fit, align, and insert. 
Instead, TaMeSo-bot uses tactile cues to retrieve and execute appropriate actions from past demonstrations, implicitly recognizing subtask transitions. This approach mirrors how humans unconsciously sense subtask transitions through touch without explicitly recognizing each contact phase~\cite{johansson2009coding}.
The integration of softness is also crucial: with rigid robots, slight pose errors during contact generate large contact forces that trigger protective stops, while the wrist's softness passively accommodates misalignment, enabling safe contact exploration and stable tactile data collection.
The wrist's softness also protects the tactile sensing system by absorbing excessive contact forces during insertion. Combined with a durable distributed tactile sensor~\cite{contcatile} that withstands the high shear forces, our system achieves reliable tactile sensing under diverse contact conditions.

While reusing prior experiences has shown substantial potential for robot manipulation~\cite{di2024effectiveness, pmlr-v229-guzey23a, yu2024mimictouch}, 
the challenge lies in learning rich representations. 
Specifically, the representation must enable effective matching of current observations to relevant past experiences based on subtle differences in sensor information, without relying on explicit state labels.
Integrating \rev{tactile cues} with other modalities such as force-torque (F/T) and proprioceptive signals is helpful for capturing rich contextual information to achieve robust manipulation.
However, existing methods
have not directly addressed 
the complex spatiotemporal correlations between tactile responses and other modalities, including robot actions. They typically process modalities separately, rely on simple concatenation~\cite{yu2024mimictouch}, or ignore temporal correlations across modalities~\cite{chen2023visuo, pmlr-v229-guzey23a}, failing to capture these spatiotemporal dependencies critical for robust contact manipulation.

\begin{figure*}[t]
    \centering
    \includegraphics[width=0.9\linewidth]{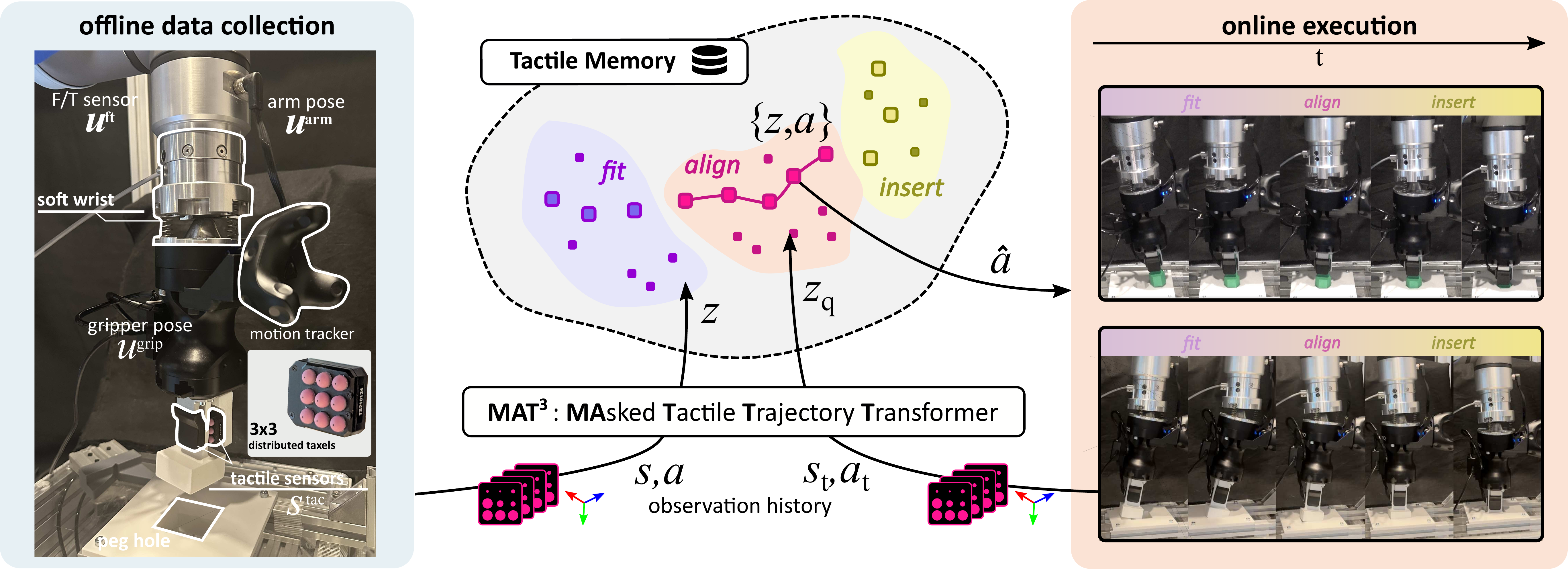}
    \caption{Overview of the \textbf{TaMeSo-bot} system. The tactile memory system stores and retrieves tactile information from a database of encoded demonstrations, enabling robust peg-in-hole manipulation by matching current sensory inputs to similar past experiences.}
    \label{fig:teaser}
\end{figure*}

To address this challenge, we design the
\textbf{Ma}sked \textbf{T}actile \textbf{T}rajectory \textbf{T}ransformer (\textbf{MAT$^3$}) that encodes spatiotemporal interaction between tactile signals and other modalities for effective storage and retrieval in tactile memory. 
\cref{fig:teaser} illustrates an overview of our TaMeSo-bot equipped with \mattt~as its core component.

Inspired by neurophysiological findings that \emph{``tactile memory encodes spatiotemporal information and interacts with other sensory modalities''}~\cite{gallace2009cognitive}, we employ a distributed tactile sensor integrated into the gripper and a built-in F/T sensor on the robot.
\mattt \,takes sequences of tokens derived from each taxel, action, and other modalities, including force/torque signals, soft wrist poses, and end-effector poses as inputs.
Through spatial and temporal encoding, the bidirectional Transformer computes cross-modal attention across different taxel locations and time horizons, enabling the extraction of spatiotemporal features that capture the intricate relationships between distributed \rev{tactile cues} and multimodal sensory signals.

To further obtain a rich feature representation relevant to the task, we train our model using masked token prediction techniques~\cite{wu2023masked,devlin-etal-2019-bert}. 
The masking strategy forces the model to reconstruct masked sensor values and actions from neighboring spatiotemporal context, compelling the model to focus on the critical contact features.
By learning from incomplete input sequences, \mattt~develops rich internal representations that capture bidirectional context dependencies for robust peg-in-hole tasks.
These learned representations are stored in the tactile memory and subsequently used via retrieve-and-replay execution, enabled by the soft wrist.

We validate our approach through real-world robot experiments on peg-in-hole tasks with seven different peg shapes and various perturbation conditions, including pose uncertainty and increased friction. 
Within the TaMeSo-bot framework,
we evaluate \mattt~against other tactile encoders and demonstrate that
\mattt~achieves higher success rates than the baselines,
handles uncertainty in grasp pose, and successfully adapts to previously unseen scenarios.
In summary, we make two key contributions: (1) \textbf{TaMeSo-bot}, which combines soft robotics with tactile memory for robust contact-rich peg-in-hole manipulation and (2) \textbf{\mattt}, a masked bidirectional Transformer encoder that extracts spatiotemporal features from distributed taxels and other modalities.

\section{Related Work}
\label{sec:related_work}

\subsection{Soft Robots and Tactile Sensors for Contact}
Soft robots equipped with tactile sensors can safely interact with objects and infer their properties~\cite{qu2023recent}, and are therefore widely used for grasping and other contact-rich manipulation tasks. Soft fingers and palms with tactile sensing have demonstrated stable grasping~\cite{lepora2021towards, shorthose2022design, zhou2022sensory, lu2024dexitac, zhang2025soft}. Whisker-inspired soft sensors support tactile exploration and object localization~\cite{lin2022whisker, sofla2024haptic}. Tendon-driven silicone grippers can estimate contact forces from tendon tension measurements and have been applied to contact-rich hair manipulation~\cite{yoo2025soft}. Pneumatic grippers with tactile sensing have been used to manipulate thin and deformable objects~\cite{zhao2025learning}.
This study tackles peg-in-hole tasks that require stable contact and high-precision manipulation. 
\final{Prior work has established that passive compliance alone enables tight-tolerance peg-in-hole insertion without tactile sensing in the control loop, though such approaches require task-specific tuning~\cite{zhang2023compliant} or heuristic segmentation~\cite{royo2023learning, azulay2025visuotactile}}. We present a retrieval-based method that leverages distributed tactile cues for zero-shot generalization to unseen peg geometries without per-task parameter adjustment.

\subsection{Tactile Representation Learning for Manipulation}
Tactile sensing captures the fine-grained interaction forces between robots and their environments that underpin dexterous manipulation~\cite{li2020review}. Tactile representation learning methods have been developed to extract essential features from spatiotemporal data obtained through vision-based or distributed tactile sensors.
Spatial structure in high-resolution tactile arrays is typically encoded with CNNs \cite{pmlr-v229-guzey23a, yu2024mimictouch}, Vision Transformer (ViT)-style Transformers \cite{higuerasparsh,chen2023visuo,zhao2024transferable}, or graph convolutions over taxel grids \cite{garcia2019tactilegcn,yang2023tacgnn,kulkarni2024tactile}.
To add \emph{temporal} information, these spatial encoders are often paired with recurrent layers such as LSTMs that process the frame-wise embeddings in sequence~\cite{dong2021tactile,ueno2024multi}.
These approaches, however, model spatial and temporal aspects separately rather than capturing their joint spatiotemporal correlations.
In this work, we propose to learn a spatiotemporal tactile representation with a bidirectional Transformer that enables (1) jointly modeling spatiotemporal correlations between distributed tactile signals and concurrent actions, (2) learning rich representations focused on critical features through masked token prediction, and (3) incorporating auxiliary modalities through weighted soft-concatenation that provides tactile-centric modality fusion.

\subsection{Retrieval from Offline Demonstrations}
A growing body of work explores retrieval-enhanced learning to control, which leverages large, unlabeled demonstration databases to retrieve relevant past experiences for training policies~\cite{du2023behavior,nasiriany2022sailor,kuroki2024iros} or decision-making~\cite{humphreys2022largescale}. This paradigm has shown promising results while eliminating the need for explicit task segmentation or manual annotation.
In parallel, another line of work~\cite{yu2024mimictouch,papagiannis2025rplusx,di2024effectiveness} pursues a non-parametric control policy via retrieve-and-replay of demonstrations without learning a parametric policy. While effective in some scenarios, some methods still require learned alignment modules or residual policies to compensate for the mismatch between retrieved demonstrations and real-world execution, adding system complexity.
In contrast to prior work, \mattt\ does not require learned alignment modules due to its spatiotemporal structure and masking strategy, enabling robust action retrieval.

In summary, by combining spatiotemporal tactile representation learning with inherent physical adaptability, our control pipeline offers a lightweight and practical alternative to existing retrieval-based frameworks for peg-in-hole tasks.

\section{TaMeSo-bot}
\label{sec:method}
This section details the core components of the tactile memory with soft robot (TaMeSo-bot).
\subsection{Overview}
 \cref{fig:teaser} provides an overview. 
 Our objective is to explore the synergistic potential of a tactile memory system with a soft wrist for robust contact-rich manipulation, validated through peg-in-hole tasks.
 We employ a rigid robot arm with an integrated F/T sensor, a soft wrist~\cite{von2020compact} and a two-finger gripper with a distributed tactile sensor (a $3\times3$ array of taxels)~\cite{contcatile}, each measuring a 3D force. The distributed tactile sensor supplies rich and localized contact information that is particularly valuable during peg-in-hole insertion. We assume access to the gripper pose via a motion capture system.
 
The soft wrist consists of three coil springs in parallel and provides a large 6D deformation, which is essential for safe data collection: a rigid arm alone tends to trigger protective stops upon contact. 
Mechanical compliance enables safe data collection and execution by absorbing large contact forces.

We collect offline demonstrations using this system via teleoperation. We then train \mattt, incorporating spatiotemporal representations from multimodal sensor streams with a Transformer encoder augmented by masked token prediction (\cref{sec:representation_learning}), and subsequently construct and query a database for retrieval-based action selection (\cref{sec:database_retrieval}).

\subsection{\mattt:~Masked Tactile Trajectory Transformer}
\label{sec:representation_learning}
The tactile memory system implements a non-parametric control policy that retrieves the appropriate action by querying a database whose keys are tactile representations augmented with multimodal contexts. To build these query keys, we aim to learn an encoder $\mathcal E$ that maps a sub-trajectory $\tau = \left( \left\lbrace \bm{s}_{t-H+1}, \bm{a}_{t-H+1} \right\rbrace, \dots, \left\lbrace \bm{s}_t, \bm{a}_t \right\rbrace \right) $
consisting of state-action pairs $\lbrace \bm s, \bm a \rbrace$ over a history window $H$, to a compact representation vector $\bm{z}_t$ that is later stored in a database with the corresponding action $\bm{a}_t$. Each embedding~$\bm z_t$ is stored in the database together with its accompanying action~$\bm a_t$ and later used to retrieve the appropriate action at execution time.

\begin{figure*}[t]
    \centering
    \includegraphics[width=\linewidth]{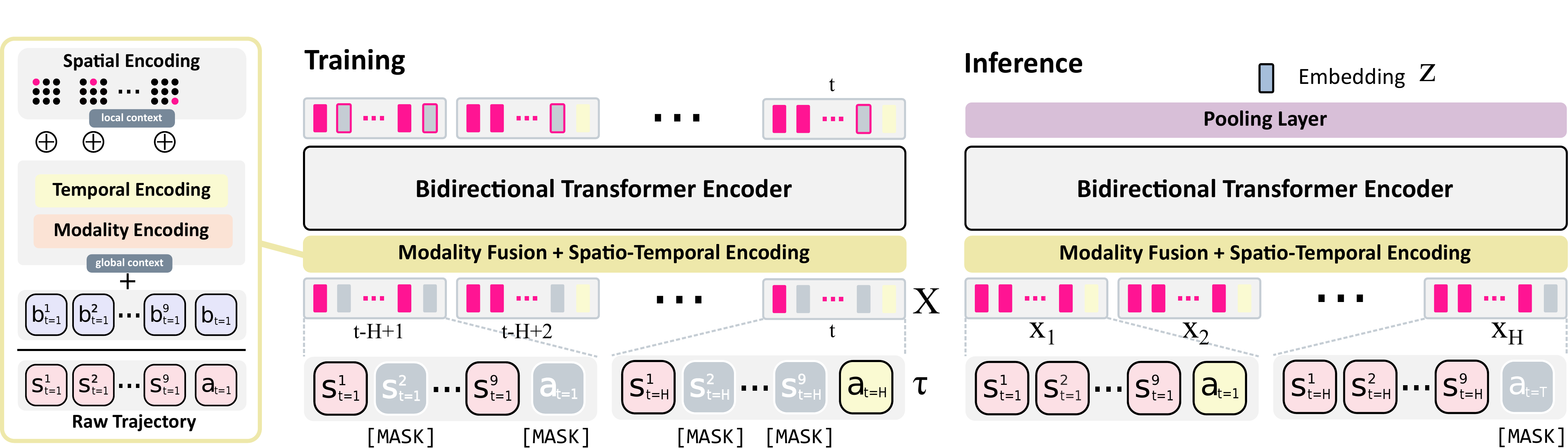}
    \caption{\textbf{Masked Tactile Trajectory Transformer (\mattt)} for distributed taxels.
    We integrate auxiliary and spatiotemporal information through soft and hard concatenation of the embeddings.
    During training, the encoder learns to reconstruct the states and actions within a time window $H$, while randomly masking input tokens. After training, the tactile trajectory datasets are encoded into a single representative embedding $\bf z$ that captures the spatiotemporal dynamics of the tactile-action sequences.}
    \label{fig:method}
\end{figure*}

\paragraph{Modality Fusion and Spatiotemporal Encoding}
At each timestep $t$, 
we collect tactile signals from a PapillArray Tactile Sensor~\cite{contcatile} with  
nine ($3\times 3$) taxels, each producing 3D force, denoted as $\bm S_t^{\mathrm{tac}} = \lbrace \bm{s}_t^{1}, \dots, \bm{s}_t^{9} \rbrace \in \mathbb{R}^{3 \times 9}$.
As auxiliary states, we capture the F/T measurements $\bm u_t^{\mathrm{ft}} \in \mathbb R^6$, arm pose $\bm u_t^{\mathrm{arm}} \in \mathbb R^7$, and gripper pose $\bm u_t^{\mathrm{grip}} \in \mathbb R^7$ measured from a motion capture system. 
We also record the robot action $\bm{a}_t \in \mathbb{R}^{6}$, which is the displacement of the 3D position $\left [\Delta x, \Delta y, \Delta z \right ]$ and Euler rotation representations $\left [ \Delta \theta_x, \Delta \theta_y, \Delta \theta_z \right ]$ of the robot arm.

\cref{fig:method} left summarizes how we integrate each modality and spatiotemporal information into an input embedding.
The individual taxels and action serve as \textbf{base tokens} $\lbrace \bm s^{1}, \dots, \bm s^{9}, \bm a \rbrace $, while the other sensory information $\lbrace \bm u^{\mathrm{ft}}, \bm u^{\mathrm{arm}}, \bm u^{\mathrm{grip}}\rbrace$ serve as \textbf{auxiliary tokens} that compensate for global contexts. These tokens are projected into a set of $d_e$-dimensional embeddings 
$\bm b_t = \lbrace \bm b^{1}, \dots, \bm b^{9}, \bm b^{\mathrm 10} \rbrace$ 
and $\bm e^{\text{aux}}_t = \lbrace \bm e^{\mathrm{ft}}_t, \bm e^{\mathrm{arm}}_t, \bm e^{\mathrm{grip}}_t\rbrace$ by a linear projection $f(\cdot)$.
We employ sinusoidal encoding for spatiotemporal modeling.
Spatial relationships between taxels are captured by assigning position embeddings
$\bm y^i \in \mathbb R^{d_{\text{pos}}}$ to each taxel at $3\times 3$ grid coordinates $(r_i, c_i) \in \{-1, 0, 1\}^2$ and concatenated to the respective tactile embeddings $\bm b^i$.
To inform temporal positions across multiple timesteps, temporal position is encoded as $\bm e^{\mathrm{time}}_t \in \mathbb R^{d_e}$.
The action token is considered at the center of this grid $(0,0)$ to facilitate spatial interaction between actions and surrounding \rev{tactile cues}.
To incorporate these auxiliary embeddings $\bm e^{\text{aux}}$ as global context, we add their weighted sum to each base token embedding $\bm b$.
In summary, each input embedding $\bm x^i \in \mathbb R^{{d_e + d_{\mathrm {pos}}}}$
integrates information through a two-level fusion of global and local context, $\bm x^i_t = \left [\bm b^i_t + \bm e^{\text{aux}}_t \right ] \oplus \bm y^i$, where
\begin{equation}
\begin{aligned}
\!\!\bm e^{\text{aux}}_t = w^{\mathrm{ft}} \bm e^{\mathrm{ft}}_t + w^{\mathrm{arm}} \bm e^{\mathrm{arm}}_t + w^{\mathrm{grip}} \bm e^{\mathrm{grip}}_t + w^{\mathrm{time}} \bm e^{\mathrm{time}}_t\,.
\end{aligned}
\label{eq:embedding}
\end{equation}
$\oplus$ denotes hard-concatenation of positional embeddings and $+$ denotes soft-concatenation with weights $w$ for auxiliary modalities and temporal information.

\paragraph{Bidirectional Transformer-based Tactile Encoder}
We introduce a Bidirectional Transformer-based tactile representation learning approach that captures spatiotemporal patterns across sensory inputs and actions.
Our model processes sub-trajectory sequences $\tau$ from historical observations. Through the embedding process defined in \cref{eq:embedding}, a sub-trajectory $\tau$ is transformed into an input tensor \\$\bm X = \left ( \lbrace \bm x^1_1, \dots, \bm x^{10}_1 \rbrace, \dots, \lbrace \bm x^1_H, \dots, \bm x^{10}_H \rbrace  \right )\in \mathbb R^{10 \times H \times d}$, where $H$ is the sequence length, 10 represents the number of tokens per timestep consisting of nine taxel tokens and one action token, and $d = d_{e} + d_{\mathrm{pos}}$ is the embedding dimension.
We employ a bidirectional transformer encoder $\mathcal E$ with $L$ layers to process the input tensor $\bm X$. Each layer applies multi-head self-attention followed by a feed-forward network, enabling the model to learn complex relationships between taxels, actions, and auxiliary modalities across the time window $H$.
This architecture effectively captures the spatiotemporal structure of tactile information while maintaining computational efficiency through the attention mechanism's ability to model long-range dependencies.
Details of the model parameters are provided in the Appendix.

\paragraph{Masked Tactile Trajectory Encoder}
To learn tactile representations that capture the intrinsic structure of the data,
we employ a masked token prediction strategy~\cite{wu2023masked, devlin-etal-2019-bert}.
As illustrated in \cref{fig:method}, we train a Bidirectional Transformer-based encoder that learns to reconstruct input states and actions with a temporal window $H$ while randomly masking input tokens.
Each modality is first passed through a dedicated linear layer that projects its raw shape to the Transformer's common embedding dimension \(d\). 
During training, a portion of the input sequence is randomly masked as \texttt{Masked}$(\bm{X})$, forcing the model to
infer missing values from neighboring spatial and temporal context.
This masking strategy compels the model to identify essential features by learning critical patterns for reconstruction, while developing context-sensitive representations across different modalities.

Specifically, for each input $\bm X$, we sample a masking ratio uniformly from the range $[0,0.6]$, following prior work~\cite{wu2023masked}.
During inference, since the current action $\bm a_t$ is unknown, we mask the action token in the current timestep and use the encoder to generate query vectors that capture the current tactile context. These learned representations are then stored in the tactile memory database and used for action retrieval during execution.

\paragraph{Learning Objective}
After encoding, each processed token is passed through a modality-specific linear layer that projects the embedding back to the original signal dimension.
The reconstructed outputs are then compared with the raw observations at the same temporal indices.  
The model is trained to reconstruct these observations by minimizing a mean-squared-error (MSE) loss averaged over every element in the entire mini-batch:

\begin{equation}
\label{eq:loss}
    \mathcal{L} = \mathcal{L}_{\text{tactile}} + \mathcal{L}_{\text{action}}\,,
\end{equation}
where $\mathcal{L}_{tactile}$ and $\mathcal{L}_{action}$ denote the MSE losses for tactile and action reconstruction, respectively, each averaged over all timesteps in the mini-batch. 
The loss in~\cref{eq:loss} is optimized with the Adam optimizer.

\paragraph{Feature Pooling}
To obtain a global tactile representation that captures information across all tokens, we apply a pooling function $g(\cdot)$.
Generally, symmetric functions serve as ideal pooling operators since they are invariant to the ordering of input tokens and can extract universal features regardless of permutation.
For our system, we adopt the widely used average pooling~\cite{lee2019set}.

\subsection{Database and Retrieval}
\label{sec:database_retrieval}
The tactile memory database stores pairs of tactile representations and their corresponding actions $\left\lbrace \bm z_t, \bm a_t \right\rbrace$, enabling the robot to retrieve appropriate behaviors during execution based on the current sensory state.

\paragraph{Database Construction}
Once an encoder $\mathcal{E}$ is pretrained by offline demonstrations, we construct the database using the collected trajectories. Each trajectory $\tau$ consists of a sequence of tactile tokens $\mathbf{x}_t \in \mathbb{R}^{10 \times H \times d}$, where $10$ denotes the number of tokens including nine tactile sensor taxels and one action token, $H$ is the number of timesteps to encode, and $d$ is the feature dimension.
To obtain a global tactile representation that is invariant to the input token ordering, we perform feature pooling using a symmetric function. Following prior work \cite{wu2023masked}, we adopt average pooling as our pooling strategy. This operation converts the input tensor $\mathbf{x}_t$ into a representative vector $\mathbf{z}_t \in \mathbb{R}^d$.
The resulting pair $\left\lbrace \mathbf{z}_t, \mathbf{a}_t \right\rbrace $, where $\mathbf{a}_t$ is the corresponding action, is then stored in the database $\mathcal{D}$.

\paragraph{Non-parametric Control Policy}
One core hypothesis is that appropriate actions can be obtained through similarity-based retrieval from past experiences, effectively transforming the tactile memory database into a non-parametric control policy $\pi : \bm s \mapsto \bm a$.
At execution time, given a sequence of tactile observations $\left ( \bm s_{t-H+1}, \dots, \bm s_{t} \right )$ and previous actions $\left (\bm a_{t-H+1}, \dots, \bm a_{t-1} \right )$, we encode it using the pretrained encoder $\mathcal E$ and pooling function $g(\cdot)$ to obtain the query representation $\mathbf{z}_q \in \mathbb{R}^d_e$. 
Here we mask the current action token as it is unknown at the time of execution. The bidirectional attention aggregates information from the remaining tokens so that $\bm z_{q}$ still captures the full tactile context.
We then retrieve the $k$ most similar representations from the database using L2 distance,

\begin{equation}
\mathrm{dist}(\bm{z}_q, \bm{z}_i) =  \| \bm z_q - \bm z_i \|_2\,.
\end{equation}
For real-time robot control at $50$ Hz, we employ an approximate nearest neighbor search using a hierarchical navigable small world (HNSW) graph-based index~\cite{malkov2018efficient}
, which offers logarithmic-time complexity while maintaining high retrieval accuracy. From the $k$ retrieved neighbors, we uniformly sample one.
The integration of masked trajectory modeling with nearest neighbor retrieval creates a robust policy that can handle partially observed contexts and adapt to novel situations by leveraging the most relevant past experiences. Furthermore, this non-parametric approach preserves the safety characteristics of demonstration data, as the robot's actions are bounded by the behaviors present in the offline dataset.

\section{Experiments}

\label{sec:experiments}
We conduct real robot experiments to evaluate our TaMeSo-bot system in peg-in-hole tasks. 
Our experiments aim to answer the following two questions: (1) Can TaMeSo-bot complete peg-in-hole tasks? and (2) Does \mattt~improve robustness and adapt to unseen pegs and conditions?

\subsection{Setup}
\paragraph{Robot}
We use a robotic arm (UR5e, Universal Robots A/S, Denmark) with an integrated F/T sensor, a parallel gripper (Hand-e, Robotiq, Canada), and a soft wrist~\cite{von2020compact}, which consists of three coil springs in parallel and allows 6D deformation. The spring constant of the spring is 1.448 N/mm, and the equilibrium length is 25 mm. A motion tracker (HTC VIVE Tracker, HTC Corporation, Taiwan) is mounted on the gripper to capture its pose, and a tactile sensor (PapillArray Tactile Sensor~\cite{contcatile}, Contactile Pty Ltd, Australia) is attached to one side of the gripper.

\begin{figure}[t]
\vspace{5pt}
    \centering
    \includegraphics[width=1\linewidth]{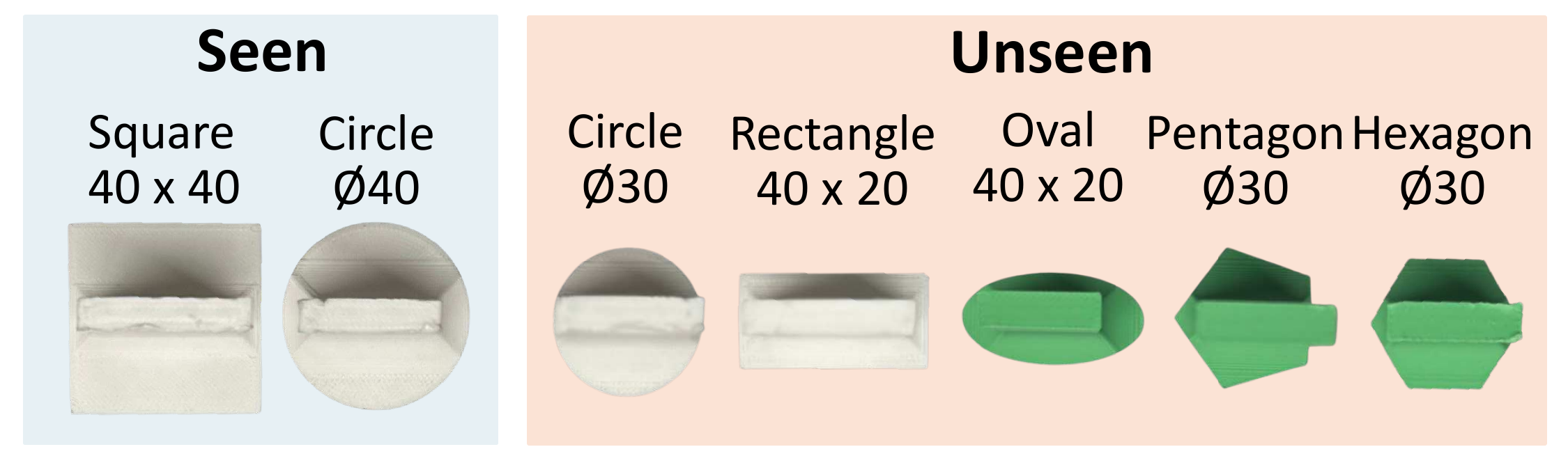}
    \caption{Seen and unseen pegs and dimensions (mm)}
    \label{fig:all-shapes-top-view}
\end{figure}

\paragraph{Pegs and Dataset Collection}
In the peg-in-hole tasks, we use seven 3D-printed pegs, including two for the training set and five for the validation set, as illustrated in \cref{fig:all-shapes-top-view}.
We collect demonstrations using the teleoperation system with a VR controller (HTC VIVE Controller, HTC Corporation, Taiwan). The robot grasps each peg in a vertical orientation. Both the sampling and control frequencies are set to 50 Hz. 
Each demonstration starts from one of four initial poses distributed equally around the hole and terminates when the peg is fully inserted into the hole, lasting approximately 150--300 time steps (3-6 seconds). The resulting dataset comprises 64 successful demonstrations: 32 from the square peg and 32 from the cylinder peg.
The tolerances of the pegs and holes are 2 mm.

\paragraph{Baselines}

Since no existing representation learning methods apply directly to our setting, which includes distributed tactile sensors, arm and gripper poses, and an F/T sensor, we compare our approach against the following baselines:

\begin{description}[leftmargin=0pt]
    \item[Tactile Transformer] We develop a Tactile Transformer by adapting the Visuo-Tactile Transformer (VTT)~\cite{chen2023visuo} to our modalities. Whereas VTT uses patched RGB images as input tokens, our Tactile Transformer uses taxels. At each timestep, each taxel, action and additional modality are encoded as separate tokens and fed to a Transformer encoder with cross-modal attention. Following the original VTT, we use learned 1D position embeddings added to the tokens, whereas \mattt~uses fixed sinusoidal 2D spatial embeddings concatenated with content embeddings. The final representation is obtained by average pooling over the output tokens. Unlike \mattt, which applies soft concatenation across modalities and incorporates temporal dependencies, the Tactile Transformer maintains separate tokens per modality.
    \item[\mattt~w/o Mask] This variant is identical to \mattt~but trained without the masking strategy, isolating the contribution of masked modeling.
\end{description}

\begin{figure}[t]
    \centering
    \includegraphics[width=1\linewidth]{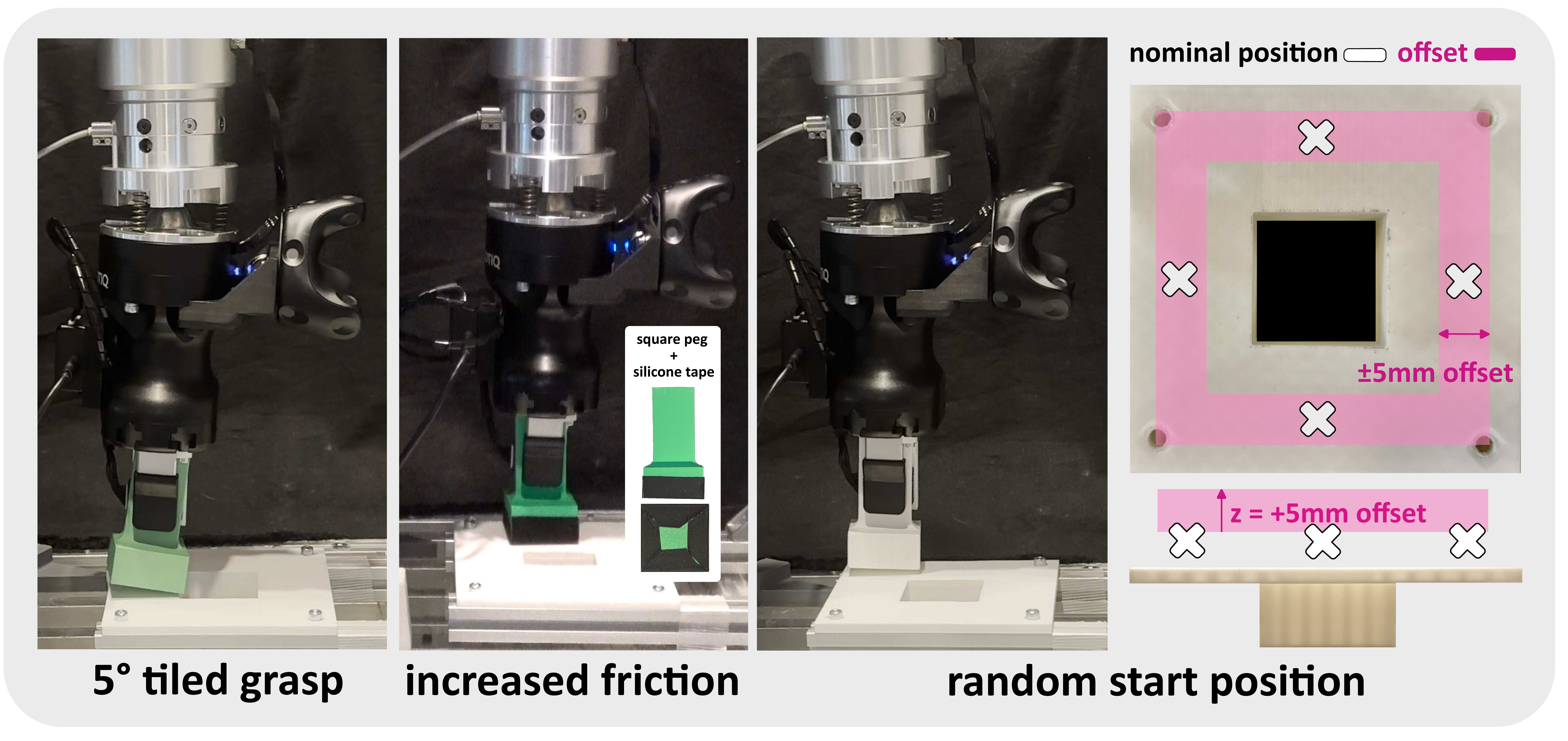}
    \caption{Unseen conditions in the experiments.}
    \label{fig:unseen_conditions}
\end{figure}

\paragraph{Unseen Conditions}
\label{sec:real-robot-evaluation}
We evaluate task success rates under both seen and unseen pegs shown in \cref{fig:all-shapes-top-view}, as well as robustness to three types of unseen perturbations illustrated in \cref{fig:unseen_conditions}: (1) different starting positions, (2) increased friction between the peg and fixture surface, and (3) tilted grasping pose. To randomize the starting position, we uniformly sample spatial offsets from the nominal starting pose used during teleoperation. To increase friction, we attach rubber tape to the peg bottom. For grasping perturbations, we manually tilt the peg by $5^\circ$ before task execution.

\paragraph{Parameter Settings}
We implement \mattt~as a $4$-layer Transformer with $8$ attention heads and hidden dimension $512$. 
We use $248$-dimensional token embeddings $d_e$ with $8$-dimensional spatial embedding $d_\text{pos}$, yielding $256$-dimensional input embedding. 
Following previous work~\cite{wu2023masked}, we sample the masking ratio for input tokens from a uniform distribution over the interval $[0, 0.6]$.
The model is trained for $30$ epochs with a batch size of $128$ and a learning rate of $1.0 \times 10^{-4}$.
\paragraph{Hyperparameter Selection}
These parameters were selected based on preliminary experiments, where we systematically evaluated different configurations. Specifically, we tested various embedding and hidden size pairs from $\{(32,64), (64,128), (128,256), (256,512), (512,1024)\}$, with $(256,512)$ yielding the best action retrieval accuracy.

\begin{table}[t]
    \caption{Real-robot evaluation in \textbf{seen and unseen pegs}. Peg-in-hole task success rate (\%) under four different starting positions. \hcb{Best} and \hcs{second-best} results are highlighted. %
    }
    \centering
    \footnotesize
    \setlength{\tabcolsep}{3pt}
    \begin{tabular}{@{}lccc|c@{}}
    \toprule
    \textbf{Shape} & \makecell{\textbf{Tactile}\\\textbf{Transformer}} & \makecell{\textbf{\mattt}\\\textbf{w/o Mask}} & \textbf{\mattt}  & \textbf{ACT~\cite{act}}\\
    \midrule
    \multicolumn{5}{@{}l@{}}{\textit{Seen}} \\
    \quad Square & 8/40 & \hcb{36/40} & \hcs{35/40} & 13/40 \\
    \quad Cyl. ($\diameter$40) & 10/40 & \hcs{36/40} & \hcs{36/40} & \hcb{38/40} \\
    \midrule
    \textbf{Total (Seen)} & \makecell{18/80 \\ (22.5\%)} & \makecell{\hcb{72/80}\\\hcb{(90\%)}} & \makecell{\hcs{71/80}\\ \hcs{(88.8\%)}} & \makecell{51/80 \\ (63.75\%)}\\
    \hline\hline 
    \multicolumn{5}{@{}l@{}}{\textit{Unseen}} \\
    \quad Cyl. ($\diameter$30) & 12/40 & 33/40 & \hcs{35/40} & \hcb{37/40} \\
    \quad Rectangle & 6/40 & \hcs{26/40} & \hcb{32/40} & 10/40 \\
    \quad Oval & 7/40 & \hcs{28/40} & \hcb{34/40} & 26/40 \\
    \quad Hexagon & 5/40 & \hcs{30/40} & \hcb{36/40} & 20/40 \\
    \quad Pentagon & 5/40 & \hcs{25/40} & \hcb{33/40} & 13/40 \\
    \midrule
    \textbf{Total (Unseen)} & \makecell{35/200 \\ (17.5\%)} & \makecell{\hcs{142/200}\\ \hcs{(71\%)}} & \makecell{\hcb{170/200}\\\hcb{(85\%)}} & \makecell{106/200 \\ (53\%)}\\
    \bottomrule
    \end{tabular}
    \label{tab:real-robot-evaluation}
\end{table}

\newcolumntype{Y}{>{\centering\arraybackslash\mdseries}X}
\newcommand{\rp}[2]{\shortstack[c]{#1\\(#2)}}  

\begin{table}[t]
    \caption{Real-robot evaluation with \textbf{unseen conditions} testing different starting positions, increased friction between the peg and the fixture surface, and a tilted grasping pose.}
    \centering
    \footnotesize
    \setlength{\tabcolsep}{3pt}
    \begin{tabular}{@{}lccc|c@{}}
    \toprule
    \textbf{Condition} & \makecell{\textbf{Tactile}\\\textbf{Transformer}} & \makecell{\textbf{\mattt}\\\textbf{w/o Mask}} & \textbf{\mattt} & \textbf{ACT}~\cite{act}\\
    \midrule
    \makecell[l]{Unseen Starting\\Positions} & 2/40 & \hcs{9/40} & \hcb{20/40} & 4/40 \\
    Increased Friction & 2/40 & \hcs{7/40} & \hcb{20/40}  & 3/40 \\
    $5^\circ$ Tilted Grasp & 5/40 & \hcs{19/40} & \hcb{29/40} & 10/40 \\
    \midrule
    \textbf{Total} & \makecell{9/120 \\ (7.5\%)} & \makecell{\hcs{35/120}\\ \hcs{(29.2\%)}} & \makecell{\hcb{69/120}\\ \hcb{(57.5\%)}} & \makecell{17/120\\(14.17\%)} \\
    \bottomrule
    \label{tab:real-robot-evaluation-unseen}
    \end{tabular}
\end{table}

\begin{table}[t]
    \caption{\textbf{Action smoothness comparison.} 
    Each cell reports the \textbf{mean} $\pm$ \textbf{std} of per--timestep action changes, computed across rollouts of the same peg shape. 
    \textbf{Pos} is the Euclidean norm of position change and \textbf{Rot} is the angular change derived from the relative rotation. 
    For each rollout, these increments are averaged over timesteps, then aggregated (mean $\pm$ std) across all rollouts. 
    \textbf{Lower values indicate smoother motion}, while greater values indicate jerky and discontinuous action.
    \hcb{Orange} marks the lower value between Transformer and \mattt{} for each row and metric. }
    \centering
    \footnotesize
    \setlength{\tabcolsep}{3pt}
    \begin{tabular}{@{}lcccc@{}}
    \toprule
    & \multicolumn{2}{c}{\textbf{Tactile Transformer}} & \multicolumn{2}{c}{\textbf{\mattt}} \\
    \cmidrule(lr){2-3}\cmidrule(lr){4-5}
    \textbf{Shape} & \textbf{Rot [$^\circ$]$\downarrow$} & \textbf{Pos [mm]$\downarrow$} & \textbf{Rot [$^\circ$]$\downarrow$} & \textbf{Pos [mm]$\downarrow$} \\
    \midrule
    \multicolumn{5}{@{}l@{}}{\textit{Seen}} \\
    \quad Square & 0.3 $\pm$ 0.2 & 2.2 $\pm$ 1.1 & \hcb{0.2 $\pm$ 0.1} & \hcb{1.7 $\pm$ 0.6} \\
    \quad Cyl. ($\diameter$40) & 0.3 $\pm$ 0.1 & \hcb{1.8 $\pm$ 0.9} & \hcb{0.2 $\pm$ 0.2} & \hcb{1.8 $\pm$ 1.3} \\
    \midrule
    \textbf{Total (Seen)} & 0.3 $\pm$ 0.2 & 2.0 $\pm$ 1.0 & \hcb{0.2 $\pm$ 0.1} & \hcb{1.7 $\pm$ 1.0} \\
    \hline\hline
    \multicolumn{5}{@{}l@{}}{\textit{Unseen}} \\
    \quad Cyl. ($\diameter$30) & \hcb{0.2 $\pm$ 0.2} & 1.8 $\pm$ 1.6 & \hcb{0.2 $\pm$ 0.2} & \hcb{1.7 $\pm$ 1.0} \\
    \quad Rectangle & 0.4 $\pm$ 0.2 & 2.5 $\pm$ 0.9 & \hcb{0.2 $\pm$ 0.1} & \hcb{1.7 $\pm$ 0.6} \\
    \quad Oval & 0.3 $\pm$ 0.2 & 2.4 $\pm$ 0.8 & \hcb{0.2 $\pm$ 0.1} & \hcb{1.6 $\pm$ 0.5} \\
    \quad Hexagon & 0.2 $\pm$ 0.1 & 2.0 $\pm$ 0.7 & \hcb{0.1 $\pm$ 0.1} & \hcb{1.4 $\pm$ 0.8} \\
    \quad Pentagon & 0.2 $\pm$ 0.1 & 2.0 $\pm$ 0.5 & \hcb{0.1 $\pm$ 0.1} & \hcb{1.8 $\pm$ 0.6} \\
    \midrule
    \textbf{Total (Unseen)} & 0.3 $\pm$ 0.1 & 2.1 $\pm$ 0.9 & \hcb{0.2 $\pm$ 0.1} & \hcb{1.6 $\pm$ 0.7} \\
    \bottomrule
    \label{tab:action-diff}
    \end{tabular}
\end{table}

\subsection{Results}
\looseness=-1
\Cref{tab:real-robot-evaluation} presents the success rates across seen and unseen pegs, with 40 trials per peg shape (10 trials from each starting position), totaling 80 trials for seen pegs and 200 trials for unseen pegs. For seen pegs, \mattt~w/o Mask and \mattt~achieved comparable performance at 90\% and 88.8\%, respectively. On unseen pegs, \mattt~significantly outperformed \mattt~w/o Mask with 85\% versus 71\% success rate, demonstrating the effectiveness of masked modeling for generalization. The Transformer baseline achieved only 22.5\% and 17.5\% on seen and unseen pegs, respectively, indicating that single-step cross-modal attention alone is insufficient for capturing the dynamics in this contact-rich manipulation task.
\rev{We also compared against Action Chunking Transformer (ACT)~\cite{act}, a parametric policy trained on the same 64 demonstrations with tactile-proprioceptive inputs. ACT achieves 63.75\% on seen pegs but drops to 53\% on unseen shapes. The dominant failure mode is applying downward force before proper alignment, triggering protective stops. We hypothesize that retrieval-based control provides implicit phase-consistent regularization by constraining actions to the demonstration dataset, whereas parametric policies can extrapolate beyond this distribution, executing insertion-phase behaviors prematurely.}

\Cref{tab:real-robot-evaluation-unseen} evaluates robustness under three unseen perturbation conditions.
\rev{
While \mattt~w/o Mask, the Transformer baseline, and ACT degraded substantially to 29.2\%, 7.5\%, and 14.17\%,}
\mattt~maintained 57.5\% success rate across all perturbations. This demonstrates that masked spatiotemporal representation learning enables robust adaptation to unseen conditions.

In summary, TaMeSo-bot with \mattt, both with and without masking, achieves near-complete success on seen pegs. Moreover, \mattt improves robustness to unseen pegs and conditions.

\section{Discussion}
This section discusses the experimental results on the smoothness of retrieved actions and the learned embedding space, and how these relate to performance. We then discuss failure cases and outline future work.

\subsection{Does \mattt~help retrieve smoother actions?}
Retrieved actions are not necessarily smooth; abrupt motions can cause task failures. Prior work~\cite{yu2024mimictouch} found action retrieval alone insufficient, potentially due to the lack of action smoothness, and therefore added a residual policy learned online via reinforcement learning.
To assess smoothness with \mattt, \Cref{tab:action-diff} reports per--timestep differences between consecutive actions for \mattt~and the Tactile Transformer. \mattt~yields smaller differences for most peg shapes, indicating that it successfully retrieves smoother actions. The supplementary video also illustrates the abrupt motions produced by the Tactile Transformer.

\subsection{Does \mattt's embedding space capture subtask-relevant information?}
\label{sec:embedding-space-analysis}
We hypothesize that our representation encodes subtask-relevant information and that retrieving actions sequentially for each subtask improves robustness. We qualitatively analyze the structure of \mattt's high-dimensional embedding space via visualization.

We build the retrieval database from $80\%$ of the demonstrations and select representative query vectors from the remaining $20\%$. To visualize the high-dimensional structure while preserving neighborhood relations, we use a two-step procedure. First, we apply t-SNE to project offline state embeddings into a 2D space. Second, to relate high-dimensional proximity to subtasks, we overlay colormaps on the projection.

\begin{figure}[t]
\vspace{5pt}
    \centering
    \includegraphics[width=\linewidth]{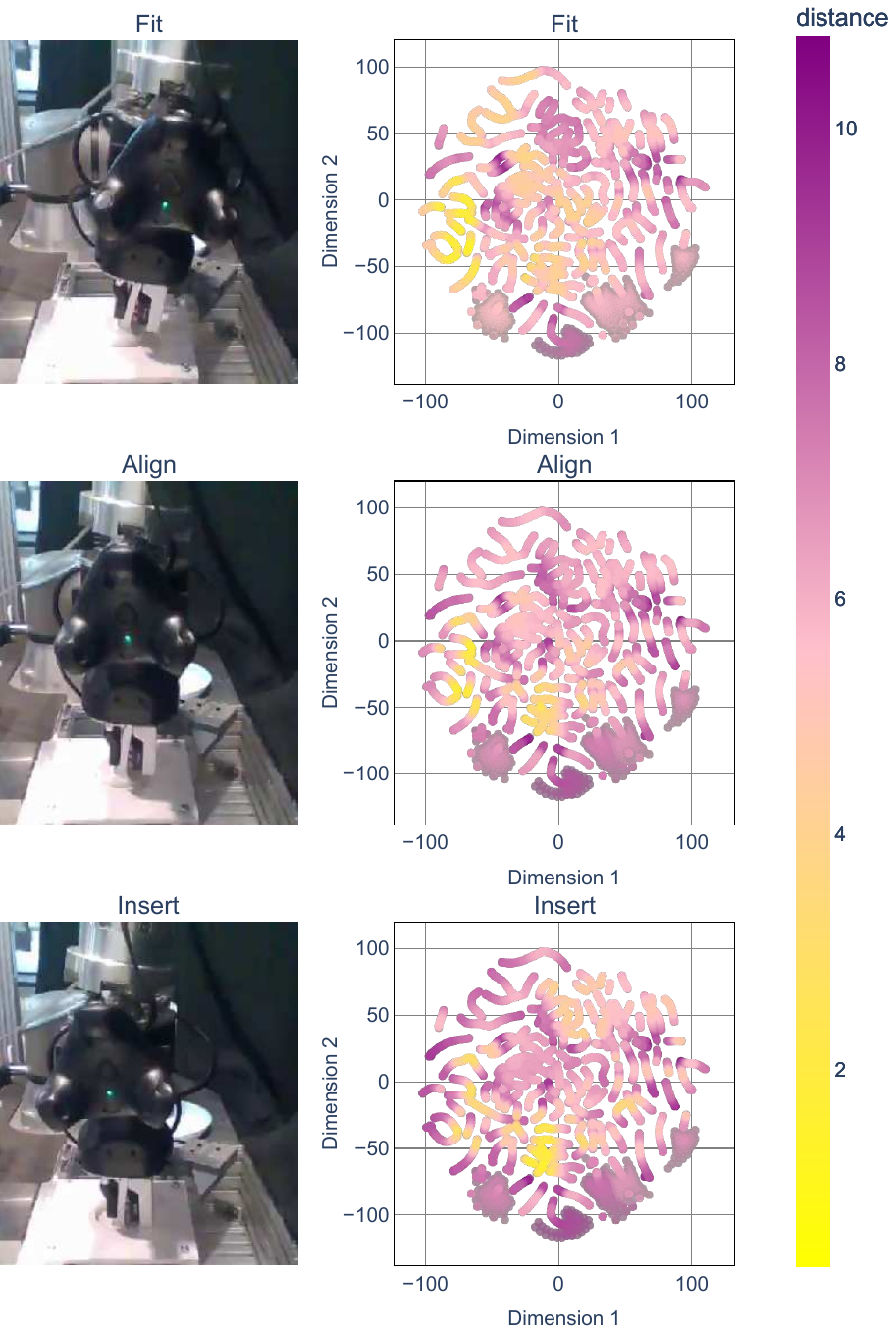}
    \caption{
        \textbf{t-SNE visualization of the learned embedding space}. 
        Colormaps indicate distance in the original high-dimensional space to representative query points for each subtask. 
        \textbf{Right}: Embedding space t-SNE visualizations of \mattt, where points are colored by L2 distance to the query vector for each corresponding subtask. \textbf{Left}: Video frames corresponding to the \emph{fit}, \emph{align}, and \emph{insert} subtasks. %
    }
    \label{fig:embedding-space-analysis}
\end{figure}

\cref{fig:embedding-space-analysis} shows colormaps of distances computed in the original embedding space between each projected state and query embeddings chosen for the subtasks \emph{fit}, \emph{align}, and \emph{insert}. Query embeddings are taken from manually segmented sequences, using states at the end of each subtask. The t-SNE plots exhibit connected yet distinct regions for \emph{fit}, \emph{align}, and \emph{insert}, indicating that \mattt\ encodes subtask-related structure, which likely contributes to robustness.

\subsection{Failure cases, limitations and future work}
Failure cases include incorrect movement direction and partial insertion, where the peg either slips past the hole or gets caught, resulting in a gradual deviation of the gripper pose.
\rev{As for limitations, this study focuses on peg-in-hole tasks; extending to other contact-rich tasks such as grinding, scooping, and wiping remains future work. Our retrieval-based approach cannot extrapolate beyond the stored distribution. Promising directions include leveraging large-scale open datasets for tactile pretraining and combining retrieval with parametric policy learning. Additionally, replacing the external motion tracker with visual pose estimation would simplify the system setup and improve reproducibility.}

\section{Conclusion}
\label{sec:conclusion}
We presented \textbf{TaMeSo-bot}, a system that integrates tactile memory with soft robotics for robust insertion tasks. Our Masked Tactile Trajectory Transformer (\textbf{\mattt}) learns spatiotemporal representations from multimodal sensor data while explicitly modeling spatial relationships across distributed taxel sensors. 
Real-robot experiments on peg-in-hole tasks demonstrate that \mattt~achieves robust generalization to unseen objects and conditions, significantly outperforming baselines without masked modeling or spatiotemporal structure. The integration of masked tactile representation learning with physical compliance enables adaptation to novel scenarios. 
We believe the synergy of tactile memory and physical compliance opens promising new research directions, potentially enabling robots to achieve human-like dexterity and adaptability in complex manipulation tasks.

\bibliographystyle{IEEEtran}

\begin{thebibliography}{10}
\providecommand{\url}[1]{#1}
\csname url@rmstyle\endcsname
\providecommand{\newblock}{\relax}
\providecommand{\bibinfo}[2]{#2}
\providecommand\BIBentrySTDinterwordspacing{\spaceskip=0pt\relax}
\providecommand\BIBentryALTinterwordstretchfactor{4}
\providecommand\BIBentryALTinterwordspacing{\spaceskip=\fontdimen2\font plus
\BIBentryALTinterwordstretchfactor\fontdimen3\font minus
  \fontdimen4\font\relax}
\providecommand\BIBforeignlanguage[2]{{%
\expandafter\ifx\csname l@#1\endcsname\relax
\typeout{** WARNING: IEEEtran.bst: No hyphenation pattern has been}%
\typeout{** loaded for the language `#1'. Using the pattern for}%
\typeout{** the default language instead.}%
\else
\language=\csname l@#1\endcsname
\fi
#2}}

\bibitem{gallace2009cognitive}
A.~Gallace and C.~Spence, ``The cognitive and neural correlates of tactile
  memory.'' \emph{Psychological Bulletin}, vol. 135, no.~3, p. 380, 2009.

\bibitem{suomalainen2022survey}
M.~Suomalainen, Y.~Karayiannidis, and V.~Kyrki, ``A survey of robot
  manipulation in contact,'' \emph{Robotics and Autonomous Systems}, vol. 156,
  p. 104224, 2022.

\bibitem{jin2021contact}
S.~Jin, X.~Zhu, C.~Wang, and M.~Tomizuka, ``Contact pose identification for
  peg-in-hole assembly under uncertainties,'' in \emph{Proc. American Control
  Conference}, 2021, pp. 48--53.

\bibitem{von2020compact}
F.~von Drigalski, K.~Tanaka, M.~Hamaya, R.~Lee, C.~Nakashima, Y.~Shibata, and
  Y.~Ijiri, ``A compact, cable-driven, activatable soft wrist with six degrees
  of freedom for assembly tasks,'' in \emph{Proc. IROS}, 2020, pp. 8752--8757.

\bibitem{johannsmeier2019framework}
L.~Johannsmeier, M.~Gerchow, and S.~Haddadin, ``A framework for robot
  manipulation: Skill formalism, meta learning and adaptive control,'' in
  \emph{Proc. ICRA}, 2019, pp. 5844--5850.

\bibitem{morgan2023towards}
A.~S. Morgan, Q.~Bateux, M.~Hao, and A.~M. Dollar, ``Towards generalized robot
  assembly through compliance-enabled contact formations,'' in \emph{Proc.
  ICRA}, 2023, pp. 8010--8016.

\bibitem{johansson2009coding}
R.~S. Johansson and J.~R. Flanagan, ``Coding and use of tactile signals from
  the fingertips in object manipulation tasks,'' \emph{Nature Reviews
  Neuroscience}, vol.~10, no.~5, pp. 345--359, 2009.

\bibitem{contcatile}
``{Contactile Accessed on: Apr. 21, 2025},'' \url{https://contactile.com/}.

\bibitem{di2024effectiveness}
N.~Di~Palo and E.~Johns, ``On the effectiveness of retrieval, alignment, and
  replay in manipulation,'' \emph{IEEE RA-L}, vol.~9, no.~3, pp. 2032--2039,
  2024.

\bibitem{pmlr-v229-guzey23a}
I.~Guzey, B.~Evans, S.~Chintala, and L.~Pinto, ``Dexterity from touch:
  Self-supervised pre-training of tactile representations with robotic play,''
  in \emph{Proc. CoRL}, 2023, pp. 3142--3166.

\bibitem{yu2024mimictouch}
K.~Yu, Y.~Han, Q.~Wang, V.~Saxena, D.~Xu, and Y.~Zhao, ``Mimictouch: Leveraging
  multi-modal human tactile demonstrations for contact-rich manipulation,'' in
  \emph{Proc. CoRL}, 2024.

\bibitem{chen2023visuo}
Y.~Chen, M.~Van~der Merwe, A.~Sipos, and N.~Fazeli, ``Visuo-tactile
  transformers for manipulation,'' in \emph{Proc. CoRL}, 2023, pp. 2026--2040.

\bibitem{wu2023masked}
P.~Wu, A.~Majumdar, K.~Stone, Y.~Lin, I.~Mordatch, P.~Abbeel, and
  A.~Rajeswaran, ``Masked trajectory models for prediction, representation, and
  control,'' in \emph{Proc. ICML}, 2023, pp. 37\,607--37\,623.

\bibitem{devlin-etal-2019-bert}
J.~Devlin, M.-W. Chang, K.~Lee, and K.~Toutanova, ``{BERT}: Pre-training of
  deep bidirectional transformers for language understanding,'' in \emph{Proc
  NAACL}, 2019, pp. 4171--4186.

\bibitem{qu2023recent}
J.~Qu, B.~Mao, Z.~Li, Y.~Xu, K.~Zhou, X.~Cao, Q.~Fan, M.~Xu, B.~Liang, H.~Liu,
  \emph{et~al.}, ``Recent progress in advanced tactile sensing technologies for
  soft grippers,'' \emph{Advanced Functional Materials}, vol.~33, no.~41, p.
  2306249, 2023.

\bibitem{lepora2021towards}
N.~F. Lepora, C.~Ford, A.~Stinchcombe, A.~Brown, J.~Lloyd, M.~G. Catalano,
  M.~Bianchi, and B.~Ward-Cherrier, ``Towards integrated tactile sensorimotor
  control in anthropomorphic soft robotic hands,'' in \emph{Proc. ICRA}, 2021,
  pp. 1622--1628.

\bibitem{shorthose2022design}
O.~Shorthose, A.~Albini, L.~He, and P.~Maiolino, ``Design of a 3d-printed soft
  robotic hand with integrated distributed tactile sensing,'' \emph{IEEE RA-L},
  vol.~7, no.~2, pp. 3945--3952, 2022.

\bibitem{zhou2022sensory}
Z.~Zhou, R.~Zuo, B.~Ying, J.~Zhu, Y.~Wang, X.~Wang, and X.~Liu, ``A sensory
  soft robotic gripper capable of learning-based object recognition and
  force-controlled grasping,'' \emph{IEEE Transactions on Automation Science
  and Engineering}, vol.~21, no.~1, pp. 844--854, 2022.

\bibitem{lu2024dexitac}
C.~Lu, K.~Tang, M.~Yang, T.~Yue, H.~Li, and N.~F. Lepora, ``Dexitac: Soft
  dexterous tactile gripping,'' \emph{IEEE/ASME Transactions on Mechatronics},
  vol.~30, no.~1, pp. 333--344, 2024.

\bibitem{zhang2025soft}
N.~Zhang, J.~Ren, Y.~Dong, X.~Yang, R.~Bian, J.~Li, G.~Gu, and X.~Zhu, ``Soft
  robotic hand with tactile palm-finger coordination,'' \emph{Nature
  Communications}, vol.~16, no.~1, p. 2395, 2025.

\bibitem{lin2022whisker}
M.~A. Lin, E.~Reyes, J.~Bohg, and M.~R. Cutkosky, ``Whisker-inspired tactile
  sensing for contact localization on robot manipulators,'' in \emph{Proc.
  IROS}, 2022, pp. 7817--7824.

\bibitem{sofla2024haptic}
M.~S. Sofla, S.~Vayakkattil, and M.~Calisti, ``Haptic localization with a soft
  whisker from moment readings at the base,'' \emph{Soft Robotics}, vol.~11,
  no.~4, pp. 539--549, 2024.

\bibitem{yoo2025soft}
U.~Yoo, N.~Dennler, E.~Xing, M.~Matari{\'c}, S.~Nikolaidis, J.~Ichnowski, and
  J.~Oh, ``Soft and compliant contact-rich hair manipulation and care,'' in
  \emph{Proc. HRI}, 2025, pp. 610--619.

\bibitem{zhao2025learning}
C.~Zhao, C.~Jiang, L.~Luo, S.~Yuan, Q.~Chen, and H.~Yu, ``Learning thin
  deformable object manipulation with a multi-sensory integrated soft hand,''
  \emph{IEEE T-RO}, vol.~41, pp. 4606--4621, 2025.

\bibitem{zhang2023compliant}
Q.~Zhang, Z.~Hu, W.~Wan, and K.~Harada, ``Compliant peg-in-hole assembly using
  a very soft wrist,'' \emph{IEEE Robotics and Automation Letters}, vol.~9,
  no.~1, pp. 17--24, 2023.

\bibitem{royo2023learning}
J.~Royo-Miquel, M.~Hamaya, C.~C. Beltran-Hernandez, and K.~Tanaka, ``Learning
  robotic assembly by leveraging physical softness and tactile sensing,'' in
  \emph{Proc. IROS}, 2023, pp. 6469--6476.

\bibitem{azulay2025visuotactile}
O.~Azulay, D.~M. Ramesh, N.~Curtis, and A.~Sintov, ``Visuotactile-based
  learning for insertion with compliant hands,'' \emph{IEEE RA-L}, vol.~10,
  no.~4, pp. 4053--4060, 2025.

\bibitem{li2020review}
Q.~Li, O.~Kroemer, Z.~Su, F.~F. Veiga, M.~Kaboli, and H.~J. Ritter, ``A review
  of tactile information: Perception and action through touch,'' \emph{IEEE
  T-RO}, vol.~36, no.~6, pp. 1619--1634, 2020.

\bibitem{higuerasparsh}
C.~Higuera, A.~Sharma, C.~K. Bodduluri, T.~Fan, P.~Lancaster, M.~Kalakrishnan,
  M.~Kaess, B.~Boots, M.~Lambeta, T.~Wu, \emph{et~al.}, ``Sparsh:
  Self-supervised touch representations for vision-based tactile sensing,'' in
  \emph{Proc. CoRL}, 2024.

\bibitem{zhao2024transferable}
J.~Zhao, Y.~Ma, L.~Wang, and E.~Adelson, ``Transferable tactile transformers
  for representation learning across diverse sensors and tasks,'' in
  \emph{Proc. CoRL}, 2024.

\bibitem{garcia2019tactilegcn}
A.~Garcia-Garcia, B.~S. Zapata-Impata, S.~Orts-Escolano, P.~Gil, and
  J.~Garcia-Rodriguez, ``Tactilegcn: A graph convolutional network for
  predicting grasp stability with tactile sensors,'' in \emph{International
  Joint Conference on Neural Networks}, 2019, pp. 1--8.

\bibitem{yang2023tacgnn}
L.~Yang, B.~Huang, Q.~Li, Y.-Y. Tsai, W.~W. Lee, C.~Song, and J.~Pan, ``Tacgnn:
  Learning tactile-based in-hand manipulation with a blind robot using
  hierarchical graph neural network,'' \emph{IEEE RA-L}, vol.~8, no.~6, pp.
  3605--3612, 2023.

\bibitem{kulkarni2024tactile}
S.~Kulkarni, S.~Funabashi, A.~Schmitz, T.~Ogata, and S.~Sugano, ``Tactile
  object property recognition using geometrical graph edge features and
  multi-thread graph convolutional network,'' \emph{IEEE RA-L}, vol.~9, no.~4,
  pp. 3894--3901, 2024.

\bibitem{dong2021tactile}
S.~Dong, D.~K. Jha, D.~Romeres, S.~Kim, D.~Nikovski, and A.~Rodriguez,
  ``Tactile-rl for insertion: Generalization to objects of unknown geometry,''
  in \emph{Proc. ICRA}, 2021, pp. 6437--6443.

\bibitem{ueno2024multi}
T.~Ueno, S.~Funabashi, H.~Ito, A.~Schmitz, S.~Kulkarni, T.~Ogata, and
  S.~Sugano, ``Multi-fingered dragging of unknown objects and orientations
  using distributed tactile information through vision-transformer and
  {LSTM},'' in \emph{Proc. IROS}, 2024, pp. 7445--7452.

\bibitem{du2023behavior}
M.~Du, S.~Nair, D.~Sadigh, and C.~Finn, ``Behavior retrieval: Few-shot
  imitation learning by querying unlabeled datasets,'' \emph{RSS}, 2023.

\bibitem{nasiriany2022sailor}
S.~Nasiriany, T.~Gao, A.~Mandlekar, and Y.~Zhu, ``Learning and retrieval from
  prior data for skill-based imitation learning,'' in \emph{Proc. CoRL}, 2023,
  pp. 2181--2204.

\bibitem{kuroki2024iros}
S.~Kuroki, M.~Nishiura, and T.~Kozuno, ``Multi-agent behavior retrieval:
  Retrieval-augmented policy training for cooperative push manipulation by
  mobile robots,'' in \emph{Proc. IROS}, 2024.

\bibitem{humphreys2022largescale}
P.~C. Humphreys, A.~Guez, O.~Tieleman, L.~Sifre, T.~Weber, and T.~P. Lillicrap,
  ``Large-scale retrieval for reinforcement learning,'' in \emph{Proc.
  NeurIPS}, A.~H. Oh, A.~Agarwal, D.~Belgrave, and K.~Cho, Eds., 2022.

\bibitem{papagiannis2025rplusx}
G.~Papagiannis, N.~Di~Palo, P.~Vitiello, and E.~Johns, ``{R+X}: Retrieval and
  execution from everyday human videos,'' in \emph{Proc. ICRA}, 2025, pp.
  8284--8290.

\bibitem{lee2019set}
J.~Lee, Y.~Lee, J.~Kim, A.~Kosiorek, S.~Choi, and Y.~W. Teh, ``Set transformer:
  A framework for attention-based permutation-invariant neural networks,'' in
  \emph{Proc. ICML}, 2019, pp. 3744--3753.

\bibitem{malkov2018efficient}
Y.~A. Malkov and D.~A. Yashunin, ``Efficient and robust approximate nearest
  neighbor search using hierarchical navigable small world graphs,'' \emph{IEEE
  TPAMI}, vol.~42, no.~4, pp. 824--836, 2018.

\bibitem{act}
T.~Zhao, V.~Kumar, S.~Levine, and C.~Finn, ``Learning fine-grained bimanual
  manipulation with low-cost hardware,'' \emph{RSS}, 2023.

\end{thebibliography}

\end{document}